\documentclass[10pt,twocolumn,letterpaper]{article}

\usepackage[pagenumbers]{iccv} 

\usepackage{graphicx}
\usepackage{booktabs}
\usepackage{wrapfig}

\usepackage{hyperref}

\usepackage{orcidlink}

\newcommand{\mask}{\texttt{[MASK]}}
\newcommand{\method}{\textsc{VISAGE}\xspace}

\usepackage{enumitem}
\usepackage{algorithm,algorithmicx,algpseudocode}
\usepackage[most]{tcolorbox}
\definecolor{pbAccent}{RGB}{0,112,128}     
\definecolor{pbHeader}{RGB}{245,245,247}   
\definecolor{pbBack}{RGB}{250,250,250}     
\definecolor{pbFrame}{RGB}{200,200,200}    

\definecolor{iccvblue}{rgb}{0.21,0.49,0.74}

\def\paperID{*****} 
\def\confName{ICCV}
\def\confYear{2025}

\title{Seeing to Ground: Visual Attention for Hallucination-Resilient MDLLMs}

\author{Vishal Narnaware\thanks{Equally contributing first author}
\quad
Animesh Gupta\footnotemark[1]
\quad
Kevin Zhai
\quad
Zhenyi Wang
\quad
Mubarak Shah\\
Institute of Artificial Intelligence, University of Central Florida
}
\begin{document}
\maketitle
\begin{abstract}
Multimodal Diffusion Large Language Models (MDLLMs) achieve high-concurrency generation through parallel masked decoding, yet the architectures remain prone to multimodal hallucinations. This structural vulnerability stems from an algorithmic flaw: the decoder ranks candidate tokens based on textual likelihood without verifying localized visual support. We establish that this language-only ranking induces an objective mismatch, where language probability mass acts as a misspecified proxy for the intended multimodal task. Consequently, we reinterpret hallucination as a localized optimization error, a phenomenon where the decoder exploits language shortcuts to maximize a proxy score at the expense of visual grounding. To address this objective mismatch, we introduce VISAGE, a training-free decoding framework that calibrates the objective at inference time. VISAGE estimates the proxy discrepancy by quantifying the spatial entropy of cross-attention distributions. By enforcing a localization consensus across attention heads, the method penalizes spatially uniform distributions and re-ranks token commitments to favor visually grounded outcomes. We provide an analytical stability guarantee establishing that VISAGE maintains a bounded objective loss under estimation error. Evaluations across hallucination-sensitive and general-purpose benchmarks demonstrate the robustness of the framework, yielding relative gains of 8.59\% on MMMU-val and 7.75\% on HallusionBench.
\end{abstract}

\section{Introduction}
\label{sec:introduction}

\begin{figure*}[t]
    \centering
    \includegraphics[width=0.9\textwidth]{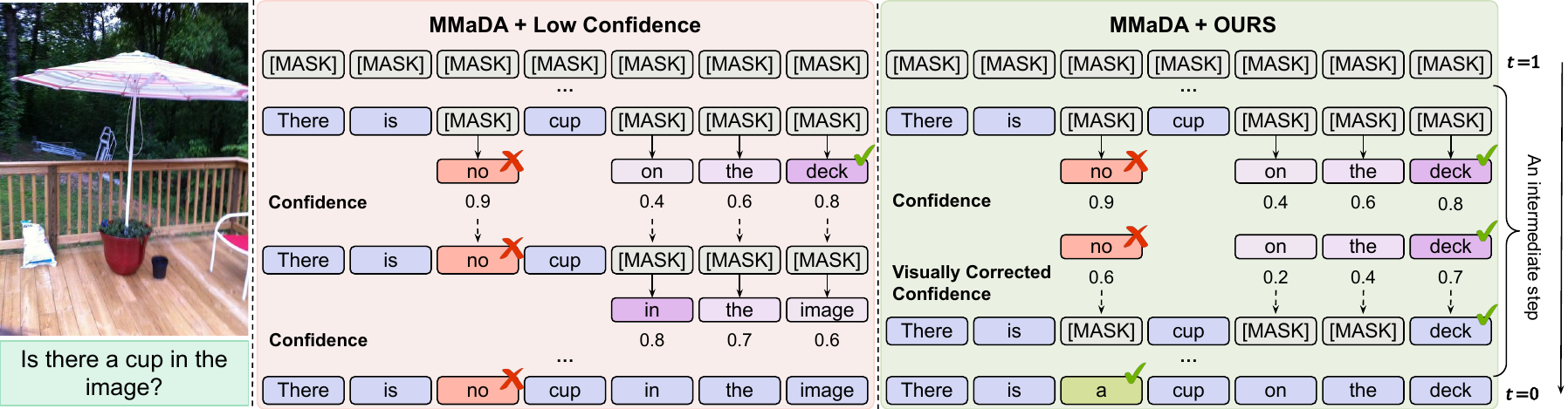}
    \caption{\textbf{VISAGE resolves language shortcuts by correcting the objective mismatch during parallel unmasking.} Given the query ``\textit{Is there a cup in the image?}'', the standard decoder (left) assigns a high language-only confidence ($0.9$) to the statistically plausible token ``\textit{no}.'' Because this proxy score lacks visual verification, the token is finalized prematurely and induces a hallucination. In contrast, VISAGE (right) calculates a visually corrected confidence ($0.6$) by estimating the proxy discrepancy $b_i$. This modification to the ranking score prevents the early commitment of the ungrounded token, allowing the model to recover the correct multimodal outcome: ``\textit{There is a cup on the deck.}''}
    \label{fig:figure1}
\end{figure*}

Multimodal Diffusion Large Language Models (MDLLMs) \cite{yang2025mmada, li2025lavida, yu2025dimple, xin2025lumina} represent a high-concurrency alternative to autoregressive decoding \cite{bai2025qwen3, wang2025internvl3, team2024gemma} for vision-language generation. By enabling parallel token generation, these architectures reduce inference latency. However, the architectural efficiency does not address the persistent problem of hallucination: the generation of responses that lack grounding in the visual input. Prior approaches often treat these generative failures as a consequence of insufficient model capacity or poor multimodal alignment during pre-training~\cite{kalai2025language}. In this work, we propose a different perspective: in the parallel unmasking setting, hallucination is a localized optimization error induced by the decoding algorithm itself. 

This localized optimization error arises because the parallel unmasking mechanism acts as an implicit, position-wise optimization process. By committing tokens based on maximum language likelihood under the denoising distribution, the decoder optimizes a language-only proxy objective rather than the intended multimodal objective. This algorithmic reliance induces an objective mismatch. Consequently, the decoder maximizes the proxy score by exploiting language shortcuts \cite{goyal2017making}, relying on textual priors to finalize predictions without verifying localized visual probability mass. As illustrated in Figure~\ref{fig:figure1} (left), when answering ``Is there a cup in the image?'', standard masked decoding assigns high probability mass to the statistically plausible token ``no''. Because the algorithm commits tokens based on the language-only proxy ranking, the ungrounded token is prematurely finalized, propagating as incorrect context for all subsequent parallel decoding steps.

\begin{figure}[t]
    \centering
    \begin{subfigure}{0.48\columnwidth}
        \centering
        \includegraphics[width=\linewidth]{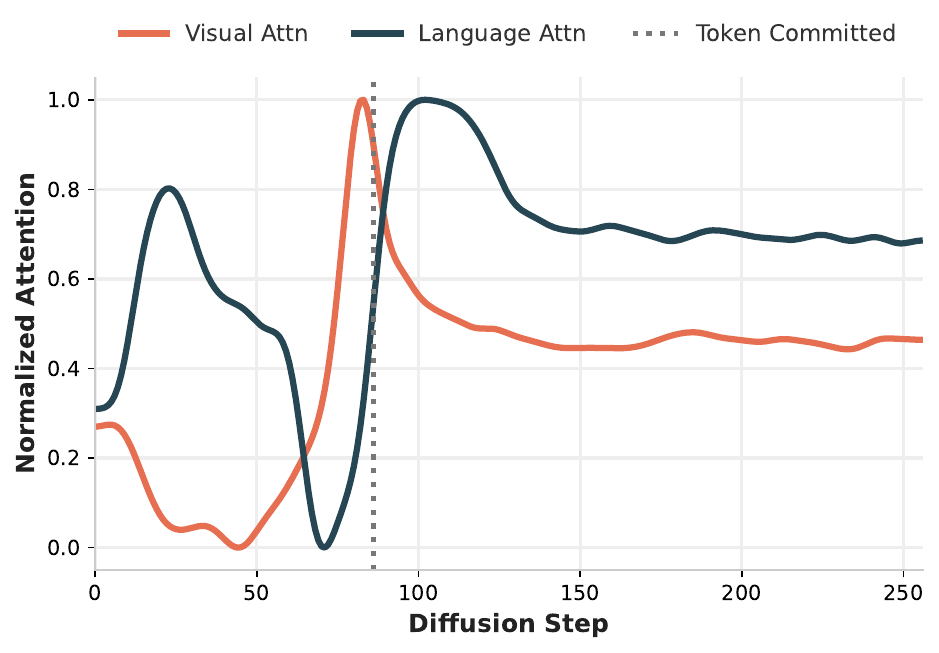}
        \caption{Grounded Token}
        \label{fig:grounded_attn}
    \end{subfigure}
    \hfill
    \begin{subfigure}{0.48\columnwidth}
        \centering
        \includegraphics[width=\linewidth]{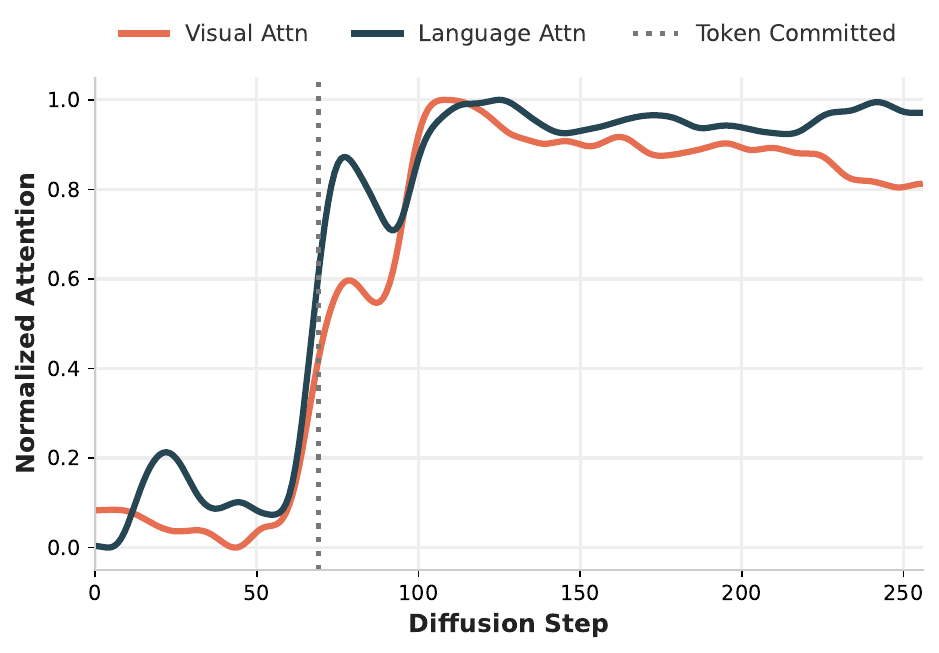}
        \caption{Hallucinated Token}
        \label{fig:hallucinated_attn}
    \end{subfigure}

    \caption{\textbf{The disparity between visual and linguistic probability mass across decoding steps suggests objective misspecification.} We visualize the normalized peak probability mass for visual (orange) and language (blue) components across the refinement process. (a) In \textbf{Visually Grounded Generation}, peak visual probability mass exceeds the language attention prior before the commitment step (vertical dotted line), consistent with a localized, low-entropy distribution over image tokens. (b) In a \textbf{Language Shortcut Hallucination}, the visual component retains a uniform spatial distribution, resulting in a suppressed peak that falls below the language attention prior. This structural disparity demonstrates that the decoder optimizes for textual likelihood while bypassing localized visual probability mass, suggesting that the spatial Shannon entropy of cross-attention provides a detectable signal for re-ranking ungrounded commitments.}
    \label{fig:figure4}
    \vspace{-4mm}
\end{figure}

To empirically validate the language shortcut hypothesis, we analyze the normalized peak probability mass for visual and language components across the diffusion steps leading up to commitment. As shown in Figure~\ref{fig:figure4}, comparing the probability mass of grounded versus hallucinated token commitments reveals a structural disparity. When the model commits a visually grounded token (left), peak visual probability mass exceeds the language prior leading up to the commitment step. In contrast, during a language shortcut hallucination (right), the visual probability mass remains spatially uniform, resulting in a suppressed peak relative to the textual context throughout the decoding process. This structural disparity demonstrates that standard probability-based ranking optimizes for textual likelihood while bypassing localized visual evidence, necessitating a re-ranking intervention at decoding time.

To resolve the objective mismatch, we introduce \method (\underline{Vis}ual \underline{A}ttention \underline{G}rounding \underline{E}ntropy), a training-free re-ranking framework. \method intercepts the parallel commitment schedule by deriving a visual re-ranking penalty from the spatial Shannon entropy of cross-attention over image tokens. The spatial entropy computation quantifies the probability mass for localized visual evidence. To ensure robustness, \method aggregates the spatial entropy across attention heads using a discrete $\beta$-quantile function. This aggregation operator functions as a localization consensus, ensuring that a token is only deemed grounded if a specific fraction of heads independently exhibit a low-entropy distribution. Tokens exhibiting spatially uniform attention distributions receive a re-ranking penalty and are downweighted, while tokens with a verified consensus of localized visual support retain their original probability mass. As illustrated in Figure~\ref{fig:figure1} (right), substituting the flawed proxy score with the entropy-calibrated probability mass penalizes shortcut-driven tokens and aligns the parallel decoding process toward grounded multimodal outcomes.

To summarize, our main contributions are as follows:
\begin{enumerate}[leftmargin=*,topsep=4pt,itemsep=4pt]
    \item \textbf{Formalization of the objective mismatch in parallel decoding.} We reinterpret hallucination in MDLLMs as a localized optimization error rather than a generative capacity failure. We establish that probability-based unmasking operates as a misspecified proxy objective that incentivizes language shortcuts.

    \item \textbf{A consensus-driven framework for grounded generation.} We introduce \method, a training-free inference framework that calibrates the decoding objective. By estimating the proxy discrepancy via the spatial entropy of cross-attention, \method enforces a localization consensus to penalize visually unsupported commitments.

    \item \textbf{Analytical stability and empirical hallucination resilience.} We establish a mathematical stability guarantee for the monotonic reweighting, proving that \method bounds the maximum objective loss under estimation error. We validate the framework across diverse multimodal benchmarks, demonstrating consistent improvements in visually grounded generation.
\end{enumerate}
\section{Related Work}

\paragraph{\textbf{Multimodal Diffusion Language Models.}}
Multimodal Diffusion Large Language Models (MDLLMs) adapt image diffusion to discrete text spaces by replacing sequential prediction with parallel masked decoding \cite{li2022diffusion, austin2021structured}. Architectures like LLaDA \cite{nie2025large} and Dream \cite{ye2025dream} improve scalability via mask-prediction objectives, while MMaDA \cite{yang2025mmada} and others \cite{li2025lavida, xin2025lumina, yu2025dimple} extend the MDLLM to joint vision-language reasoning. These high-concurrency architectures reduce inference latency compared to autoregressive counterparts. However, the discrete proposal space breaks continuous alignment mechanisms, such as classifier guidance \cite{shen2024understanding, wallace2023end, shi2023exploring, yu2023freedom} and classifier-free guidance \cite{ho2022classifier, shen2024rethinking}, because the continuous mechanisms require backpropagating perturbations through a differentiable score function. \method addresses the non-differentiable limitation by introducing a discrete re-ranking framework that recovers the grounding objective without continuous gradient approximations.

\paragraph{\textbf{Visual Grounding and Hallucination.}}
Hallucination mitigation relies on training-time alignment \cite{peters2007reinforcement, dong2023raft, deng2024prdp, wallace2024diffusion, rafailov2023direct, liang2024step} or autoregressive inference-time interventions via external verification or internal calibration. External verification pipelines like Woodpecker \cite{yin2024woodpecker} and MARINE \cite{zhao2025mitigating} utilize auxiliary object detectors to validate content post-hoc, introducing multi-pass latency. Internal calibration methods instead utilize the base model's representations to apply contrastive penalties \cite{leng2024mitigating, xu-etal-2025-mitigating}, suppress attention heads \cite{sarkar2025mitigating}, or constrain information bottlenecks \cite{zhang2026vib} during sequential next-token prediction. While these calibration interventions mitigate linguistic biases, the sequential mechanisms ignore the localized optimization error specific to parallel masked decoding, where the unmasking algorithm selects tokens based on the categorical textual distribution. The reliance on the textual distribution enables ungrounded language shortcuts \cite{goyal2017making} where language priors override visual conditioning. Unlike autoregressive interventions, \method resolves the objective mismatch by deriving a robust grounding entropy from internal attention distributions. The robust entropy ensures token commitments possess high probability mass across localized visual evidence, providing a training-free defense against decoding-induced hallucinations.

\begin{figure*}[t]
    \centering
    \includegraphics[width=0.9\textwidth]{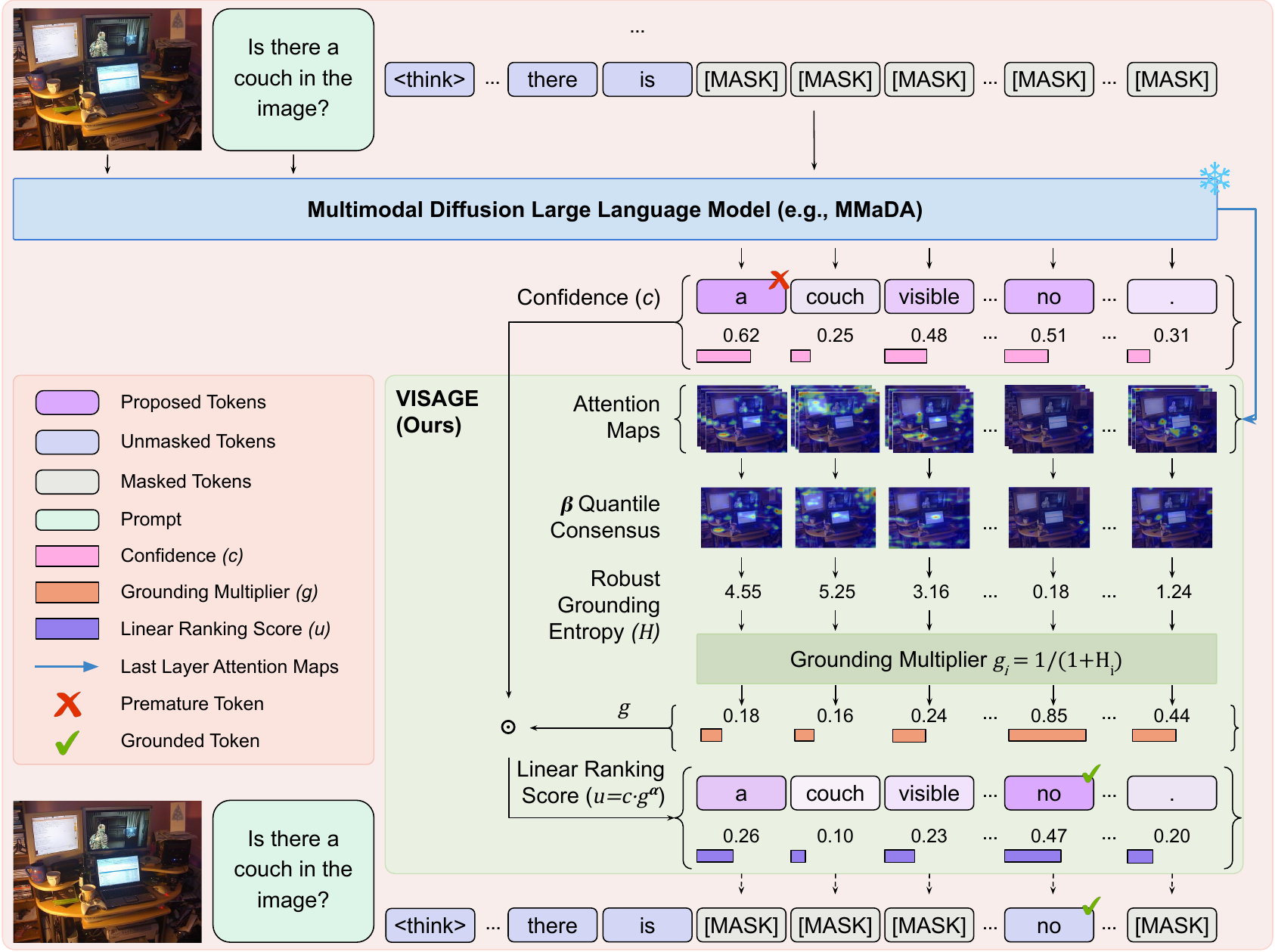}

\caption{
\textbf{Overview of VISAGE during parallel masked decoding.} At each decoding step, the frozen MDLLM generates candidate tokens alongside their initial confidence $c$. To verify visual support, we extract each candidate token's last-layer cross-attention weights over the image and compute Shannon entropy for each attention head. We then aggregate these values across heads via a $\beta$-quantile operator, yielding Robust Grounding Entropy $H$, which quantifies the concentration of visual support. A penalty multiplier $g = 1/(1+H)$ is then computed. Tokens are subsequently re-ranked using the linear ranking score $u = c \cdot g^{\alpha}$ ($\alpha=0.5$). As illustrated, the ungrounded token (\textit{``a''}) is penalized by high entropy, ensuring the visually supported token (\textit{``no''}) attains a high linear ranking score and is successfully committed to the sequence.
}

    \label{fig:figure2}
\end{figure*}

\section{Problem Formulation}
\label{sec:problem}

Let $\mathcal V$ be the vocabulary and $x$ denote the text prompt. For a visual input $v$, the intermediate MDLLM sequence at step $t\in\{1,\dots,T\}$ is represented by $y^{(t)}=[x; v; y_1^{(t)}, \dots, y_L^{(t)}]$, where each response position $y_i^{(t)} \in \mathcal V\cup \{\mask\}$. Decoding initializes all $L$ response positions as $y_i^{(1)}=\mask$ and replaces these mask tokens with discrete tokens over $T$ steps. At each decoding step $t$, the set of masked token positions is defined as:
\begin{equation}
C_t \;=\; \{\, i \in \{1,\dots,L\} : y^{(t)}_i=\mask \,\}.
\label{eq:eligible}
\end{equation}

\noindent For each masked token position $i \in C_t$, the model processes the sequence $y^{(t)}$ to compute a discrete probability distribution over the vocabulary:
\begin{equation*}
p_\theta(y_i = \nu \mid y^{(t)}, i), \qquad \forall \nu \in \mathcal V.
\end{equation*}

\noindent Let $\hat y_i \in \mathcal V$ denote the token proposal at position $i$ derived via greedy maximization of the categorical distribution. We define the position-wise confidence $c_i$ as the probability mass assigned to the proposal:
\begin{equation}
c_i \;\triangleq\; p_\theta(y_i = \hat y_i \mid y^{(t)}, i).
\label{eq:conf}
\end{equation}

\paragraph{Standard decoding optimizes a proxy objective independent of visual context.}
Governed by an unmasking schedule, the decoder is allocated a token budget $k_t$ for the current step. The decoder must select a subset of positions $U_t\subseteq C_t$ such that $|U_t|=k_t$. The sequence is then updated by committing the proposed tokens at these selected positions:
\begin{equation}
y^{(t+1)}_i \;=\; 
\begin{cases} 
\hat y_i, & i\in U_t,\\
y^{(t)}_i, & \text{otherwise.}
\end{cases}
\label{eq:update}
\end{equation}
Confidence-based ranking selects this subset by solving the following maximization problem:
\begin{equation}
U_t \triangleq \arg\max_{\substack{U\subseteq C_t\\ |U|=k_t}}
\sum_{i\in U} \log c_i.
\label{eq:std_unmask}
\end{equation}
This maximization objective is equivalent to selecting the top-$k_t$ token positions ranked by the scalar $c_i$.

The formulation in Equation~\ref{eq:std_unmask} reveals a structural vulnerability. Because Equation~\ref{eq:std_unmask} optimizes raw text probability, the equation acts as a misspecified proxy objective that omits visual grounding verification. A high confidence score $c_i$ reflects statistical language priors derived from the initial prompt or previously unmasked tokens, rather than grounding in the visual input $v$. Consequently, the unmasking objective favors the language prior and enables the model to take language shortcuts, committing tokens that lack visual support. The mismatch between the proxy objective and the intended multimodal objective necessitates a decoding-time correction mechanism capable of verifying localized visual grounding before token commitment.
\section{VISAGE: Grounding-Aware Decoding}
\label{sec:method}

To address the reliance on language-only proxy objectives that induce hallucinations, we introduce VISAGE, a decoding-time re-ranking framework. Our method formalizes hallucination as a localized optimization error under a misspecified ranking score, and introduces a cross-attention spatial entropy estimator to approximate the unobservable multimodal objective.

\subsection{Sequential Commitment under Objective Misspecification}
\label{subsec:reward_hacking}

As established in Equation~\ref{eq:std_unmask}, standard decoding ranks candidate tokens by the proxy score $r_{\mathrm{proxy}}(i) \triangleq \log c_i$. This score represents the textual likelihood of a candidate token but lacks a mechanism to quantify spatial concentration over the visual input. The intended multimodal decoding objective $r_{\star}$ incorporates both textual likelihood and localized visual correspondence:
\begin{equation}
r_{\star}(i) \;=\; \log c_i \;+\; \lambda\, r_{\mathrm{vis}}(i),
\label{eq:true_reward}
\end{equation}
where $\lambda > 0$ scales the visual grounding strength. Because a perfect oracle for visual correspondence $r_{\mathrm{vis}}$ requires computationally expensive external verification that defeats the efficiency of parallel decoding, standard algorithms default to the language-only proxy $r_{\mathrm{proxy}}$. This reliance induces an objective mismatch where $r_{\mathrm{proxy}}(i) \neq r_{\star}(i)$. We formalize this objective mismatch as the proxy discrepancy $b_i \ge 0$:
\begin{equation}
r_{\star}(i) \;=\; \log c_i \;-\; b_i.
\label{eq:discrepancy}
\end{equation}
The proxy discrepancy $b_i$ quantifies the magnitude by which textual probability mass overrides localized visual evidence. This mismatch allows the decoder to maximize the proxy score by committing tokens with high textual likelihood even when the proxy discrepancy $b_i$ is large. Because these committed tokens provide the conditioning context for subsequent steps, such ungrounded commitments propagate as incorrect context throughout the parallel decoding process. 

The deviation from the intended multimodal objective $r_{\star}$ defined in Equation~\ref{eq:true_reward} is a position-wise phenomenon captured by the proxy discrepancy $b_i$:
\begin{equation}
b_i \;=\; r_{\mathrm{proxy}}(i) - r_{\star}(i).
\label{eq:overvalue_gap}
\end{equation}
Under the discrepancy formulation in Equation~\ref{eq:overvalue_gap}, hallucinations correspond to candidate positions that exhibit high textual likelihood $r_{\mathrm{proxy}}(i)$ but suffer from a large position-wise discrepancy $b_i$. By treating hallucination as a localized optimization error rather than a global sequence failure, we apply targeted re-ranking penalties to indices where visual correspondence is insufficient.

\subsection{Grounding-Aware Objective Correction}
\label{subsec:grounding_reward}

\paragraph{\textbf{Spatial entropy quantifies the localization of visual correspondence.}}
Approximating the unobservable multimodal objective $r_\star(\cdot)$ in Equation~\ref{eq:discrepancy} requires a computable estimate of the proxy discrepancy $b_i$. We do not assume that cross-attention weights represent the exact causal influence of visual features; instead, we use the spatial concentration of normalized cross-attention score distributions as a proxy for visual evidence. As observed in the attention trajectories of Figure~\ref{fig:figure4}, grounded token commitments coincide with high peak probability mass over specific image regions, consistent with a localized, low-entropy distribution. Conversely, when the model relies on textual priors, the visual probability mass is distributed across the image tokens, resulting in a spatially uniform distribution with a suppressed peak. We formalize this relationship by deriving a computable estimator $\hat{b}_i$ from the spatial entropy of the cross-attention distribution, penalizing tokens that lack concentrated visual probability mass.

\paragraph{\textbf{Cross-attention entropy yields the grounding-aware ranking score.}}
Let $\mathcal I_{\text{img}}$ be indices of image tokens of visual input $v$ and $N:=|\mathcal{I}_{\text{img}}|$ denote the number of image tokens. At step $t$, let $A^{(t,h)}_{i,j}$ denote the last-layer cross-attention from the query representation at masked text position $i$ to image token $j \in \mathcal{I}_{\text{img}}$ in head $h\in\{1,\dots,M\}$. We renormalize attention over image tokens to obtain a normalized spatial distribution $\tilde A^{(t,h)}_{i,(\cdot)}$:
\begin{equation}
\tilde A^{(t,h)}_{i,j}
\;=\;
\frac{A^{(t,h)}_{i,j} + \frac{\delta}{N}}
{\sum_{j' \in \mathcal{I}_{\text{img}}} A^{(t,h)}_{i,j'} + \delta},
\qquad j \in \mathcal{I}_{\text{img}},
\label{eq:renorm_attention}
\end{equation}
where $\delta\!>\!0$ is a numerical smoothing constant for stability. We interpret the distribution $\tilde A^{(t,h)}_{i,(\cdot)}$ as the spatial localization of visual evidence for the candidate token $\hat y_i$. To quantify this concentration of probability mass, we define the Shannon entropy $H_i^{(t,h)}$ for the $h$-th attention head at token position $i$:
\begin{equation}
H_i^{(t,h)}
\;=\;
-\!\sum_{j \in \mathcal{I}_{\text{img}}}
\tilde A^{(t,h)}_{i,j}
\log\!\big(\tilde A^{(t,h)}_{i,j} \big).
\label{eq:head_entropy}
\end{equation}

In standard vision-language architectures, individual attention heads occasionally collapse to spurious artifacts, yielding artificially low entropy even when the token is ungrounded. To prevent a single spuriously sharp head from bypassing the grounding penalty, we aggregate the $M$ head-wise entropies into a single robust grounding entropy scalar $H_i$. We define the discrete quantile function $q_{\beta}(\cdot)$ as an operator that selects the $\lceil \beta M \rceil$-th smallest value from a set. This operator functions as a \textit{localization consensus}: rather than trusting the head with the lowest entropy, we require a specific fraction of heads to agree that a token is grounded. Let $H_{i,(1)} \le H_{i,(2)} \le \cdots \le H_{i,(M)}$ denote the head-wise entropy values $\{H_i^{(t,h)}\}_{h=1}^{M}$ sorted in ascending order. We define the aggregate robust grounding entropy $H_i$ as the specific value at index $\lceil \beta M \rceil$ in the sorted set:
\begin{equation}
H_i \;\triangleq\; q_{\beta}\big(\{H_i^{(t,h)}\}_{h=1}^{M}\big) \;=\; H_{i,(\lceil \beta M \rceil)}, \qquad \beta \in (0,1].
\label{eq:quantile_entropy}
\end{equation}
This quantile selection ensures that the aggregate entropy $H_i$ takes a numerically small value only if at least $\beta M$ individual heads also exhibit low entropy. For instance, setting $\beta\!=\!0.5$ defines $H_i$ as the median head entropy. This median requirement enforces a majority consensus: the resulting discrepancy estimator $\hat{b}_i$ remains small, thereby preserving the token's ranking, only if at least half of the attention heads independently yield concentrated probability mass over the image tokens.

We define $\hat b_i$ as the computable estimator for the true proxy discrepancy $b_i$. We convert the robust grounding entropy $H_i$ into this estimator $\hat b_i$, which increases with higher entropy:
\begin{equation}
\hat b_i \;\triangleq\; \alpha \log(1 + H_i),
\qquad \alpha \ge 0,
\label{eq:est_bias}
\end{equation}
where $\alpha$ acts as a coefficient to scale the penalty relative to the log-probability. This penalty yields the corrected log-score $\hat r(i) = \log c_i - \hat b_i$. Defining the grounding multiplier $g_i \triangleq \frac{1}{1+H_i}$, we transform the log-space score into the linear ranking score $u_i$:
\begin{equation}
u_i
\;=\;
c_i \cdot g_i^{\alpha}
\;=\; 
c_i \cdot (1+H_i)^{-\alpha}.
\label{eq:visage_score}
\end{equation}
Under the $k_t$ budget constraint, the optimal commitment set $U_t$ is the subset that maximizes the sum of corrected scores:
\begin{equation}
U_t
\;=\;
\arg\max_{\substack{U \subseteq C_t \\ |U| = k_t}}
\sum_{i \in U}
\hat r(i).
\label{eq:set_objective}
\end{equation}
Because the objective is separable across candidates, the exact optimizer is given by the top-$k_t$ positions ranked by $u_i$:
\begin{equation}
U_t
\;=\;
\operatorname{TopK}\big(\{u_i\}_{i \in C_t}, k_t\big).
\label{eq:topk_solution}
\end{equation}

\paragraph{\textbf{Bounded estimation error ensures stable and grounded commitments.}}
We provide a detailed description of the VISAGE procedure in the Supplementary Material. In practice, the entire spatial entropy derivation reduces to a highly efficient inference mechanism: VISAGE reweights token proposals monotonically via the multiplier $g_i^\alpha$. For a fixed confidence $c_i$, the ranking score $u_i$ decreases as the robust grounding entropy $H_i$ increases. This monotonic relationship ensures that the decoder commits a token only when the token is both confident and validated by localized visual correspondence. Analytically, the VISAGE framework provides a stability guarantee against estimation error between the entropy-based estimator $\hat b_i$ and the true proxy discrepancy $b_i$. If the spatial entropy estimator $\hat b_i$ approximates the proxy discrepancy $b_i$ such that the absolute error satisfies $|\hat b_i - b_i| \le \varepsilon_t$ for all candidates $i \in C_t$, the resulting objective loss is at most $2k_t\varepsilon_t$ relative to the optimal grounded subset. This $2k_t\varepsilon_t$ bound arises from the potential for a worst-case rank reversal: an overestimated suboptimal token (inflated by $+\varepsilon_t$) can displace an underestimated optimal token (deflated by $-\varepsilon_t$) in the index, resulting in a maximum per-token objective loss of $2\varepsilon_t$. We provide the formal derivation of this stability bound in the Supplementary. By directly addressing the proxy discrepancy, this bounded-error framework ensures the commitment process remains stable and visually grounded.
\section{Experiments}
\label{sec:experiments}

\paragraph{\textbf{Datasets.}} 
To evaluate hallucination mitigation, we utilize POPE~\cite{li2023evaluating} to probe basic object existence and HallusionBench~\cite{guan2024hallusionbench} to assess language-driven hallucinations and visual illusions. Furthermore, to ensure the grounding interventions do not compromise overall generation quality, we benchmark general multimodal capabilities on MMMU-val~\cite{yue2024mmmu} for college-level, multi-discipline question answering, and MME~\cite{fumme} for comprehensive vision-language tasks.

\paragraph{\textbf{Baseline Methods.}} 
We evaluate VISAGE against two primary decoding strategies. Our base model is MMaDA~\cite{yang2025mmada}, which employs standard confidence-based unmasking. Because standard decoding selects tokens based on predicted textual likelihood, standard confidence-based unmasking remains vulnerable to language shortcut bias. As a strong inference-time baseline, we adapt Visual Contrastive Decoding (VCD)~\cite{leng2024mitigating} to the masked diffusion framework. Originally designed for autoregressive models, we implement VCD by contrasting logit distributions from the original image against a visually distorted counterpart at each parallel unmasking step. This contrast penalizes reliance on linguistic priors when visual evidence is intentionally degraded. 

\paragraph{\textbf{Implementation Details.}} 
We build VISAGE decoding upon the MMaDA architecture. Our method uses a fixed spatial entropy quantile threshold $\beta=0.25$ across all experiments. For the grounding penalty, we apply a default of $\alpha=0.5$ for most visually intensive benchmarks. However, for MME, we reduce the penalty to $\alpha=0.3$. Because MME uniquely evaluates both fine-grained perception and complex cognition tasks such as commonsense reasoning and translation, this lower penalty ensures the model retains the necessary linguistic and cognitive priors without over-regularizing the visual grounding. We justify this in \Cref{sec:ablations} via ablation in \Cref{tab:ablation_alpha}. We omit restrictive post-prompts and extend the maximum generation lengths to elicit intermediate token sequences. All evaluated methods share identical generation configurations within a given benchmark for fair comparison. While method-specific hyperparameters remain constant, generation-specific parameters vary by task complexity. For POPE, HallusionBench, and MME, we set the generation length and diffusion steps to 256, with a block length of 32. For MMMU-val, we maintain a generation length of 256, but utilize 128 diffusion steps and a block length of 64.

\paragraph{\textbf{Evaluation.}} 
We evaluate model performance across the extended generation trajectories using two strategies. For POPE and MME, we use string-matching rules, comparing the final responses against the ground-truth labels to compute F1 metrics and the MME score. For HallusionBench and MMMU-val, we adopt the Qwen3-8B~\cite{yang2025qwen3} as an external LLM-as-a-judge. The LLM judge is prompted to analyze whether the generated output aligns with the ground truth.
\begin{table*}[t]
  \centering
  \small
  \setlength{\tabcolsep}{6pt}
  \renewcommand{\arraystretch}{1.15}
  \caption{\textbf{VISAGE improves visual reasoning across subject-specific and diagnostic benchmarks.}. MMaDA + VISAGE (Ours) consistently outperforms the baseline and VCD across multiple benchmarks. We observe substantial gains over the base model, notably $+2.65$ on HallusionBench and $+2.33\%$ on MMMU-val.}
  \label{tab:main_results}
  \resizebox{0.85\textwidth}{!}{
  \begin{tabular}{lcccc}
    \toprule
    \textbf{Method} &
    \shortstack{\textbf{MMMU-val} \\ (Acc \% $\uparrow$)} &
    \shortstack{\textbf{HallusionBench} \\ (Acc \% $\uparrow$)} &
    \shortstack{\textbf{POPE} \\ (F1 $\uparrow$)} &
    \shortstack{\textbf{MME} \\ (Score $\uparrow$)} \\
    \midrule
    MMaDA (Base)                & 27.11 & 34.18 & 75.97 & \textbf{1383.29} \\
    MMaDA + VCD                 & 28.44 & 34.80 & 75.85 & 1342.21 \\
    MMaDA + \method (Ours)      & \textbf{29.44} & \textbf{36.83} & \textbf{76.17} & 1372.05 \\
    \midrule
    Relative Gain \% (Ours vs. Base)   & +8.59\% & +7.75\% & +0.26\% & -0.81\% \\
    \bottomrule
  \end{tabular}
  }
\end{table*}
\section{Main Results}
\label{sec:results}

\paragraph{\textbf{Linear Ranking Score Reduces Language Shortcut Exploitation.}} We present results on hallucination and general purpose benchmarks in Table~\ref{tab:main_results}. Our method, \method, yields improvements on hallucination-sensitive benchmarks, with a +7.75\% relative gain on HallusionBench, +8.59\% on MMMU-val, and +0.26\% on POPE over the MMaDA baseline, while sustaining only a negligible relative decrease of -0.81\% on MME. The large improvement on HallusionBench reflects its unique design. The benchmark exposes models that over-rely on textual priors when faced with deceptive visual contexts, this gain validates our core hypothesis. By penalizing ungrounded tokens, \method prevents the decoder from defaulting to statistically plausible but unverified language shortcuts, forcing it to ground its generation in actual image content.

\begin{figure}
    \centering
    \includegraphics[width=0.90\linewidth]{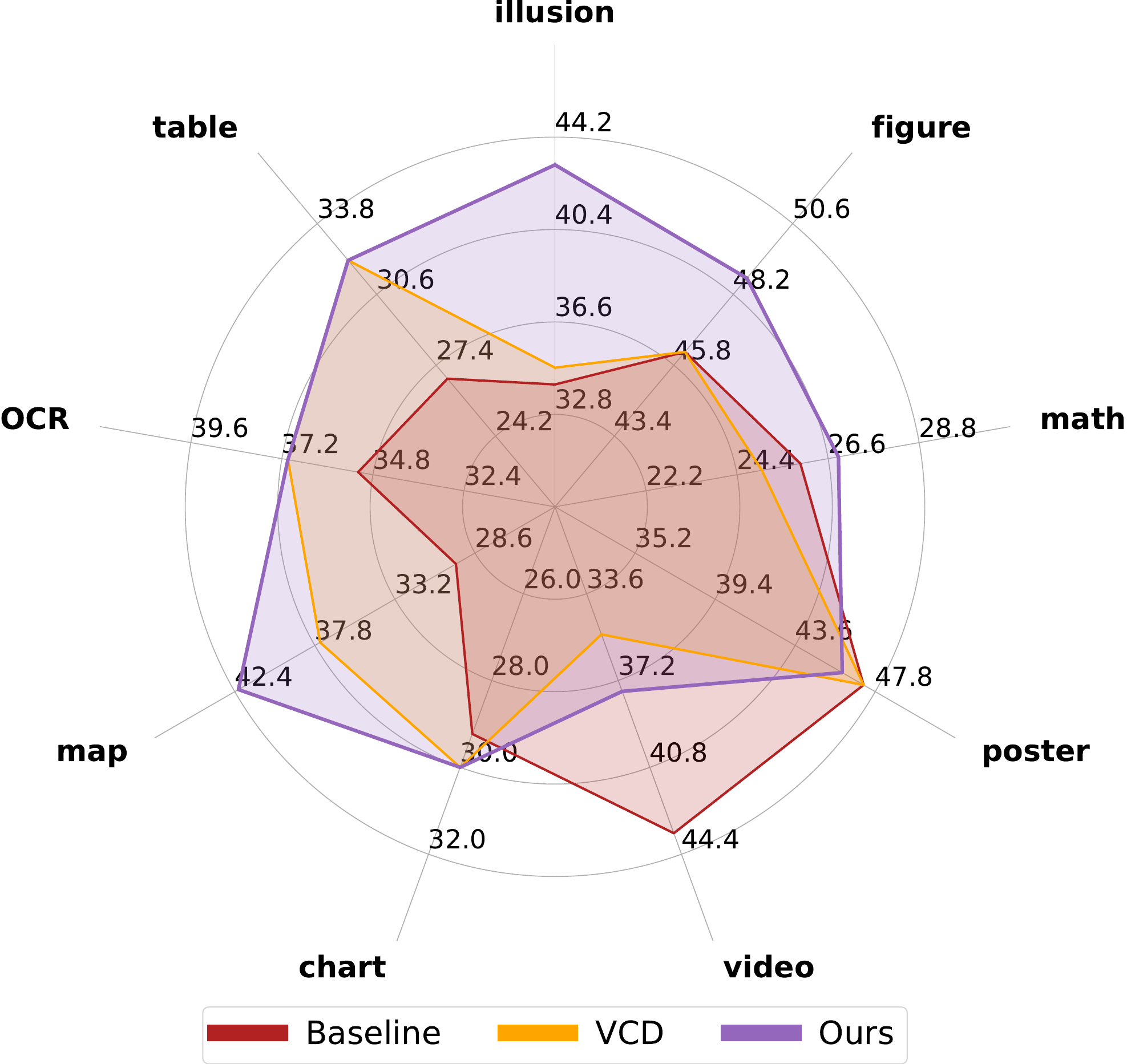}

    \caption{
        \textbf{HallusionBench Category Analysis.} Radar chart comparing \method against the Baseline and VCD. Our method achieves robust improvements across \textit{illusion} and spatial-reasoning (\textit{map}, \textit{figure}) categories.
    }
    \label{fig:hallusionbench}
    \vspace{-1em} 
\end{figure}

While POPE evaluates object-level hallucination, the relative gain (+0.26\%) is compressed by systemic label noise in the underlying dataset. Because we enable intermediate reasoning traces, \method performs exhaustive spatial verification before committing to a token. This rigorous grounding identifies valid objects that are absent from the ground-truth annotations. POPE is constructed using MS COCO \cite{lin2014microsoft}, which is known to have missing labels and false positive object assertions \cite{tkachenko2023objectlab}. Qualitative analysis reveals numerous instances where \method rightly refutes incorrect ground-truth assertions (penalized as false negatives). Consequently, these inherited label inaccuracies artificially cap the measurable improvement on this benchmark. Detailed qualitative examples of these ground-truth failures are provided in the supplementary material.

\paragraph{\textbf{Robustness to Deceptive Visual Contexts.}} Figure~\ref{fig:hallusionbench} illustrates the domains driving our overall performance gain on HallusionBench. \method substantially expands the accuracy in highly spatial subsets, such as \textit{illusion} ($\approx$9\%) and \textit{map} (12.5\%). High accuracy in these specific categories requires precise feature localization to resolve visually deceptive or complex layouts. 
Notably, VCD offers negligible improvement in the \textit{illusion} subset, suggesting that while contrastive decoding penalizes generic statistical priors, it fails to explicitly enforce the concentrated spatial verification needed to overcome deceptive visual prompts. 
Furthermore, we observe consistent improvements across dense, structured formats such as \textit{figure}, \textit{math}, and \textit{OCR}, where fine-grained semantic details are easily hallucinated by language priors. However, the baseline retains a distinct advantage in the \textit{video} category. Video reasoning requires aggregating and grounding information across multiple frames, whereas our method derives grounding from frame-level attention without explicitly modeling temporal consistency. Extending the objective correction mechanism to capture cross-frame grounding remains an important direction for future work.

\begin{figure*}[t]
    \centering
    \includegraphics[width=0.9\textwidth]{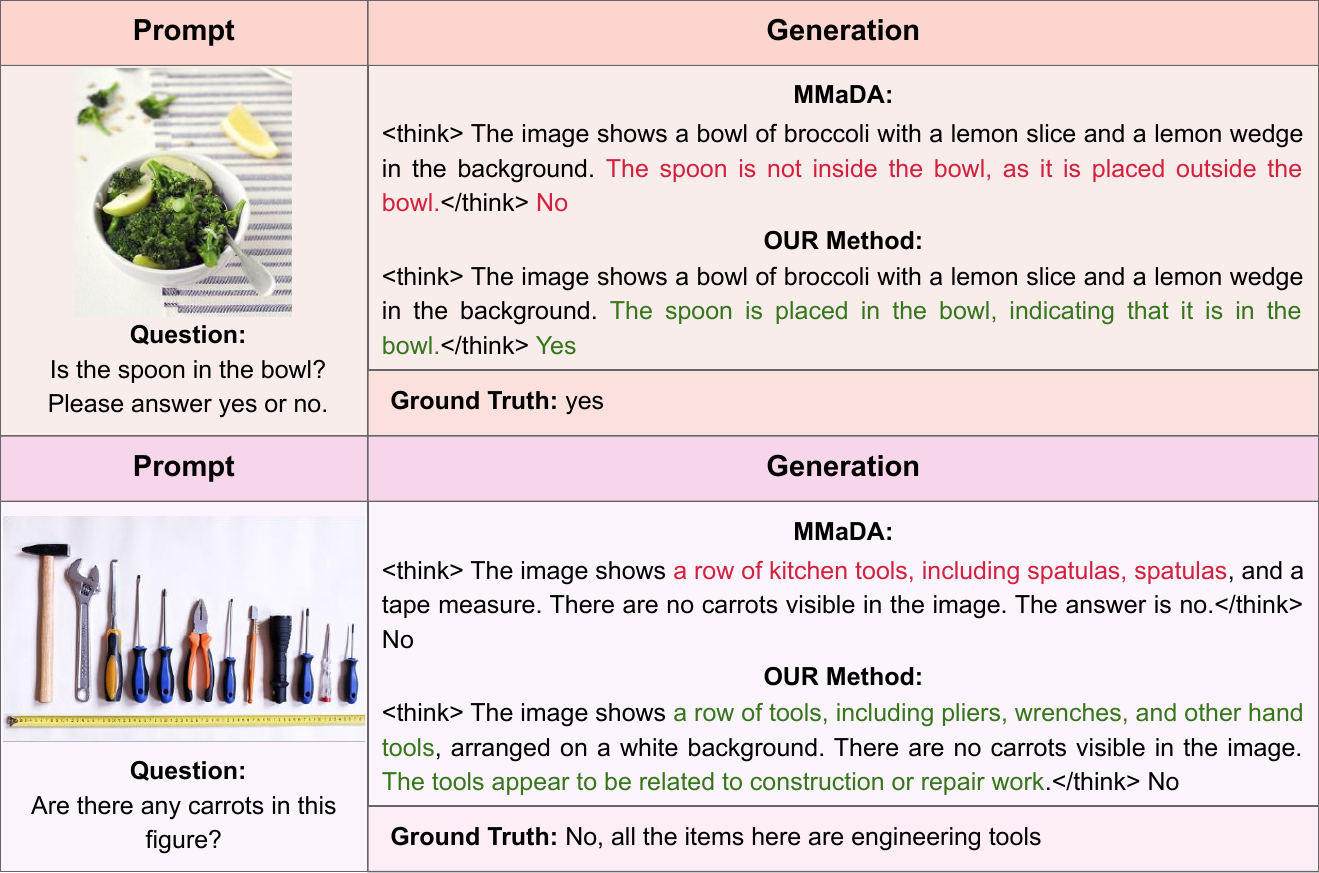}
\caption{
Qualitative comparison of visually grounded yes/no questions. We compare MMaDA and our method on two examples where hallucination arises from language shortcuts. \textbf{Top:} MMaDA incorrectly answers ``\textit{No}'' by over-relying on the textual prior that a spoon rests outside a bowl. Our method correctly grounds its ``\textit{Yes}'' decision in localized visual evidence. \textbf{Bottom:} While both methods correctly answer ``\textit{No}'', MMaDA hallucinates irrelevant objects in its intermediate thinking trace. In contrast our method maintains strictly visually consistent descriptions.
}
    \label{fig:qualitative}
    \vspace{-1em}
\end{figure*}

\paragraph{\textbf{Grounding-Aware Re-ranking Stabilizes Multi-Step Reasoning.}} \method achieves a +8.59\% relative gain on MMMU-val (Table~\ref{tab:main_results}), which requires college-level, discipline specific multi-step reasoning. In such rigorous tasks, premature commitment to an ungrounded token during intermediate reasoning steps often triggers a cascading failure, where subsequent generations condition on the hallucinated prior rather than the visual input. By re-ranking candidate tokens via our corrective penalty to the confidence, \method ensures that each step of the reasoning remains anchored to the image. This continuous regularization prevents the decoder from drifting into language shortcuts, stabilizing the generation of complex, long-form answers.

\paragraph{\textbf{Overcoming Prompt-Driven Hallucinations.}} \method neutralizes hallucinations induced by the language bias of the user's prompt. \Cref{fig:qualitative} illustrates this token-level correction. In the top example, the baseline defaults to a statistical heuristic, asserting the spoon is ``placed outside the bowl'' rather than verifying the actual spatial coordinates. Our method successfully anchors the generation to the correct visual relationship. The bottom example highlights a more severe lexical trigger: the prompt asked if there are \textit{``carrots''} in an image of engineering tools, both methods correctly conclude with ``\textit{No}.'' However, this lexical trigger (\textit{``carrots''}) causes MMaDA to undergo a language shortcut. It ignores the visual evidence of engineering tools and confidently hallucinates ``\textit{kitchen tools, including spatulas}'' to satisfy the food-related prompt. Our method suppresses this prompt-driven language shortcuts, forcing the model to correctly identify ``\textit{pliers}'' and ``\textit{wrenches}.''

\subsection{Ablative Studies.}
\label{sec:ablations}

\paragraph{\textbf{Effect of Penalty Factor $\alpha$.}} The penalty factor $\alpha$ scales the penalty applied to the language model's confidence (log-probability). It interpolates between language-driven confidence and the grounded multimodal objective, controlling how strongly visually ambiguous tokens are downweighted during commitment. 



\begin{table}
  \centering
  \setlength{\tabcolsep}{8pt}
  \renewcommand{\arraystretch}{1.15}
  \caption{Effect of $\alpha$. Sensitivity analysis of the grounding penalty on MME scores.}
  \label{tab:ablation_alpha}
  \begin{tabular}{cc}
    \toprule
    \textbf{Penalty ($\alpha$)} & \textbf{MME} (Score $\uparrow$) \\
    \midrule
    0.5  & 1320.77 \\
    0.3  & \textbf{1372.05} \\
    0.1  & 1362.68 \\
    \bottomrule
  \end{tabular}
  \vspace{-1em}
\end{table}

The ablation results in \Cref{tab:ablation_alpha} support this interpretation. On the MME benchmark, which includes cognition tasks such as Commonsense Reasoning, Text Translation, Numerical Calculation, Code Reasoning, a moderate penalty ($\alpha = 0.3$) achieves the best score (1372.05). It outperforms a stronger penalty ($\alpha = 0.5$, yielding 1320.77), suggesting that excessive penalization can degrade performance when visual grounding is not the primary source of hallucination. Conversely, applying a weak penalty ($\alpha = 0.1$, yielding 1362.68) fails to adequately suppress premature token commitments. Without sufficient regularization of the model's confidence, the decoder is not forced to seek concentrated visual evidence, leaving it vulnerable to the language shortcuts our method aims to prevent.

\paragraph{\textbf{Robustness via $\beta$-Quantile Consensus.}} Table~\ref{tab:ablation_quantile} validates the necessity of our $\beta$-quantile consensus ($q_{\beta}$) for robust grounding entropy. On the MMMU-val benchmark, the $q_{\beta}$ strategy achieves 29.44\% accuracy, outperforming both mean (28.78\%) and minimum (28.56\%) aggregation. 


\begin{table}
  \centering
  \setlength{\tabcolsep}{2pt}
  \renewcommand{\arraystretch}{1.15}
  \caption{Ablative Study on head aggregation. Quantile aggregation improves robustness over mean pooling.}
  \label{tab:ablation_quantile}
  \resizebox{0.8\linewidth}{!}{%
  \begin{tabular}{c c}
    \toprule
    \textbf{Aggregation} & \textbf{MMMU-val} (Acc\% $\uparrow$) \\
    \midrule
    Min over Heads       & 28.56 \\
    Mean over Heads      & 28.78 \\
    $q_{\beta}$ (Ours)   & \textbf{29.44 }\\
    \bottomrule
  \end{tabular}%
  }
  \vspace{-1em}
\end{table}



In multimodal transformers, individual attention heads can produce spuriously sharp distributions even when a token is not grounded. A minimum-entropy is therefore vulnerable to a single sharp head falsely signaling grounding. Conversely, mean pooling dilutes genuine localized signals if most heads remain diffuse. The $q_{\beta}$ consensus provides a principled compromise: it requires that visual grounding is consistently concentrated across a non-trivial consensus of heads before assigning grounding penalty. By filtering out both isolated artifacts and uninformative diffuse heads, $q_{\beta}$ consensus ensures stable and reliable spatial verification during token commitment.

\section{Conclusion}

We identify a structural limitation in Multimodal Diffusion Large Language Models (MDLLMs): probability-based masked decoding optimizes a language-only proxy objective, inducing hallucination through language shortcuts. By reframing parallel unmasking as a localized optimization error under objective misspecification, we establish that hallucinated tokens arise from a quantifiable proxy discrepancy rather than generative capacity failure. To resolve this objective mismatch, we introduce \method, a training-free re-ranking framework that calibrates the decoding objective at inference time. By deriving a discrepancy estimator from the spatial entropy of cross-attention distributions and enforcing a localization consensus, \method downweights tokens lacking concentrated visual probability mass. Evaluations across hallucination-sensitive and general-purpose benchmarks demonstrate the robustness of the framework. Our results establish that structural interventions on the discrete decoding objective, without parameter updates, resolve multimodal misalignments in parallel decoding architectures.

\paragraph{\textbf{Limitations and Future Work.}} VISAGE addresses visual grounding during parallel masked decoding for image-to-text generation. The current framework does not explicitly incorporate the temporal token dimensions characteristic of video-based MDLLMs. While spatial entropy generalizes to spatio-temporal attention maps, VISAGE does not enforce cross-frame temporal consistency. Extending the entropy-based consensus to explicitly model motion dynamics and temporal grounding in video-based parallel decoding represents a promising direction for future research.

\clearpage

{
    \small
    \bibliographystyle{ieeenat_fullname}
    \bibliography{references}
}

\clearpage

\appendix
\onecolumn
\setcounter{page}{1}
\renewcommand{\thefigure}{A.\arabic{figure}} %
\setcounter{figure}{0} 
\renewcommand{\thetable}{A.\arabic{table}}
\setcounter{table}{0} 
\renewcommand{\thesection}{\Alph{section}} %
\setcounter{section}{0}
\renewcommand{\theequation}{A.\arabic{equation}} %
\setcounter{equation}{0}

\newcounter{prompt}
\renewcommand{\theprompt}{P\arabic{prompt}}

\newtcolorbox{promptbox}[2][]{%
  enhanced,
  rounded corners,
  boxrule=0.5pt,
  colback=white,                 
  colframe=black!40,             
  colbacktitle=pbHeader,         
  coltitle=pbAccent,             
  fonttitle=\bfseries,
  drop shadow={opacity=0.1},     
  left=4pt, right=4pt, top=4pt, bottom=4pt,
  title={\refstepcounter{prompt}\theprompt: #2},
  #1
}

\begin{center}
  {\LARGE \bf Supplementary Material \par}
  \vspace{1em}
\end{center}

\section{Analytical Derivation of the Stability Bound}
\label{supp:proof_stability}

In this section, we provide the formal derivation bounding the objective loss of the VISAGE commitment step under estimation error. 

Let $C_t$ denote the set of candidate masked indices at decoding step $t$. Let $r_\star(i)$ be the intended multimodal objective score for candidate $i \in C_t$, and let $\hat r(i)$ be the VISAGE corrected log-score. We assume the spatial entropy estimator bounds the proxy discrepancy error such that for all $i \in C_t$:
\begin{equation}
    |\hat r(i) - r_\star(i)| \;\le\; \varepsilon_t.
    \label{eq:proof_bound_assumption}
\end{equation}

Under the budget constraint $k_t$, let $U^\star$ denote the optimal commitment set that maximizes the intended objective, and let $\hat U$ denote the set selected by VISAGE via the corrected log-score:
\begin{align}
    U^\star \;&=\; \arg\max_{\substack{U \subseteq C_t \\ |U| = k_t}} \sum_{i \in U} r_\star(i), \\
    \hat U \;&=\; \arg\max_{\substack{U \subseteq C_t \\ |U| = k_t}} \sum_{i \in U} \hat r(i).
\end{align}

We define the objective loss $\mathcal{L}$ incurred by selecting $\hat U$ instead of the optimal set $U^\star$ as:
\begin{equation}
    \mathcal{L} \;=\; \sum_{j \in U^\star} r_\star(j) \;-\; \sum_{i \in \hat U} r_\star(i).
\end{equation}

To bound the objective loss $\mathcal{L}$, we partition the sets into their intersection and disjoint components. Let $S_{\mathrm{shared}} = U^\star \cap \hat U$ denote the tokens correctly selected, $S_{\mathrm{missed}} = U^\star \setminus \hat U$ denote the optimal tokens rejected, and $S_{\mathrm{false}} = \hat U \setminus U^\star$ denote the suboptimal tokens falsely accepted. Because both $U^\star$ and $\hat U$ have exactly $k_t$ elements, the disjoint sets have equal cardinality: $|S_{\mathrm{missed}}| = |S_{\mathrm{false}}| = m$, where $m \le k_t$.

The objective loss simplifies to the difference over the disjoint sets:
\begin{equation}
    \mathcal{L} \;=\; \sum_{j \in S_{\mathrm{missed}}} r_\star(j) \;-\; \sum_{i \in S_{\mathrm{false}}} r_\star(i).
    \label{eq:proof_disjoint_loss}
\end{equation}

From the bounded error assumption in Equation~\ref{eq:proof_bound_assumption}, we bound the intended scores using the estimated scores:
\begin{align}
    r_\star(j) \;&\le\; \hat r(j) + \varepsilon_t \quad \forall j \in S_{\mathrm{missed}}, \label{eq:proof_upper} \\
    r_\star(i) \;&\ge\; \hat r(i) - \varepsilon_t \quad \forall i \in S_{\mathrm{false}}. \label{eq:proof_lower}
\end{align}

Substituting the inequalities from Equations~\ref{eq:proof_upper} and \ref{eq:proof_lower} into Equation~\ref{eq:proof_disjoint_loss} yields:
\begin{align}
    \mathcal{L} \;&\le\; \sum_{j \in S_{\mathrm{missed}}} \big(\hat r(j) + \varepsilon_t\big) \;-\; \sum_{i \in S_{\mathrm{false}}} \big(\hat r(i) - \varepsilon_t\big) \\
    \;&=\; \left( \sum_{j \in S_{\mathrm{missed}}} \hat r(j) \;-\; \sum_{i \in S_{\mathrm{false}}} \hat r(i) \right) \;+\; 2m\varepsilon_t.
    \label{eq:proof_grouped}
\end{align}

By the definition of $\hat U$, the set $\hat U$ maximizes the sum of the estimated scores $\hat r$. Therefore, the sum of the estimated scores for the selected tokens in $S_{\mathrm{false}}$ must be greater than or equal to the sum of the estimated scores for the rejected tokens in $S_{\mathrm{missed}}$:
\begin{equation}
    \sum_{i \in S_{\mathrm{false}}} \hat r(i) \;\ge\; \sum_{j \in S_{\mathrm{missed}}} \hat r(j).
\end{equation}
Consequently, the difference term $\left( \sum_{j \in S_{\mathrm{missed}}} \hat r(j) - \sum_{i \in S_{\mathrm{false}}} \hat r(i) \right)$ is less than or equal to zero. Applying this inequality to Equation~\ref{eq:proof_grouped}, we obtain:
\begin{equation}
    \mathcal{L} \;\le\; 0 \;+\; 2m\varepsilon_t.
\end{equation}
Because the number of swapped tokens $m$ is bounded by the unmasking budget $k_t$ ($m \le k_t$), the maximum objective loss is bounded:
\begin{equation}
    \mathcal{L} \;\le\; 2k_t\varepsilon_t.
\end{equation}
This derivation concludes the proof. The factor of two formally captures the worst-case rank reversal, where a suboptimal candidate is overestimated by $+\varepsilon_t$ and an optimal candidate is underestimated by $-\varepsilon_t$.

\section{Detailed VISAGE Decoding Procedure}
\label{supp:visage_algorithm}

Algorithm~\ref{alg:visage} provides the pseudo-code for \method. The algorithm illustrates the implementation within the parallel masked decoding loop. The procedure demonstrates how \method interacts with the standard unmasking schedule ($\{k_t\}_{t=1}^N$). At each step $t$, the algorithm isolates the masked candidate positions ($C_t$), extracts their corresponding last-layer cross-attention maps ($A^{(t)}$), and computes the head-wise spatial entropies. After applying the $\beta$-quantile consensus ($q_\beta$) to obtain the robust grounding entropy ($H_i$), the initial confidence ($c_i$) is monotonically re-weighted to produce the linear ranking score ($u_i$). The top-$K$ tokens ranked by $u_i$ are then committed to the sequence $Y$. Because the algorithm utilizes the internal attention distributions already computed during the forward pass of the MDLLM, \method introduces negligible computational overhead and requires no auxiliary verification models.

\begin{algorithm*}[t]
\caption{Visual Attention Grounding Entropy (VISAGE) Decoding}
\label{alg:visage}
\small
\begin{algorithmic}[1]
\Require Text instruction $X$, visual tokens $v$, answer length $L$, image indices $\mathcal I_{\text{img}}$, mask token \mask
\Require Steps $N$, unmask schedule $\{k_t\}_{t=1}^N$, penalty $\alpha\ge 0$, quantile $\beta\in(0,1]$, smoothing $\delta>0$
\State $Y \gets [X; v; \mask,\dots,\mask]$ \Comment{$L$ masks appended after text and vision}
\State $\mathcal{A}\gets\{|X|+|v|+1,\dots,|X|+|v|+L\}$ \Comment{Isolate response indices}
\For{$t=1$ to $N$}
  \State Run model on $Y$ to obtain discrete distributions $p_\theta$ and last-layer cross-attentions $A^{(t)}$
  \State $C_t \gets \{ i\in\mathcal{A} : Y_i=\mask \}$ \Comment{Identify masked candidate positions}
  \If{$C_t=\emptyset$} \State \textbf{break} \EndIf
  \ForAll{$i\in C_t$}
    \State $\hat y_i \gets \arg\max_{\nu \in \mathcal{V}} p_\theta(y_i = \nu \mid Y)$ \Comment{Greedy token proposal}
    \State $c_i \gets p_\theta(y_i = \hat y_i \mid Y)$ \Comment{Calculate initial probability mass}
    \For{$h=1$ to $M$}
      \State $Z \gets \sum_{k\in \mathcal I_{\text{img}}} A^{(t,h)}_{i,k} + \delta$
      \State $\tilde A^{(t,h)}_{i,j} \gets (A^{(t,h)}_{i,j} + \delta/|\mathcal I_{\text{img}}|)/Z \quad \forall j\in \mathcal I_{\text{img}}$
      \State $H^{(t,h)}_i \gets -\sum_{j\in \mathcal I_{\text{img}}}\tilde A^{(t,h)}_{i,j}\log(\tilde A^{(t,h)}_{i,j})$
    \EndFor
    \State $H_i \gets q_{\beta}(\{H^{(t,h)}_i\}_{h=1}^M)$ \Comment{Apply $\beta$-quantile localization consensus}
    \State $u_i \gets c_i\cdot(1+H_i)^{-\alpha}$ \Comment{Calculate linear ranking score}
  \EndFor
  \State $K\gets \min(k_t, |C_t|)$
  \State $U_t \gets \operatorname{TopK}(\{u_i\}_{i\in C_t}, K)$
  \ForAll{$i\in U_t$} \State $Y_i \gets \hat y_i$ \Comment{Commit tokens at selected indices} \EndFor
\EndFor
\State \Return $Y_{|X|+|v|+1 : |X|+|v|+L}$
\end{algorithmic}
\end{algorithm*}

\section{Qualitative Examples}
\label{supp:qualitative}

\subsection{HallusionBench}

\begin{figure*}[t]
    \centering
    \includegraphics[width=\textwidth]{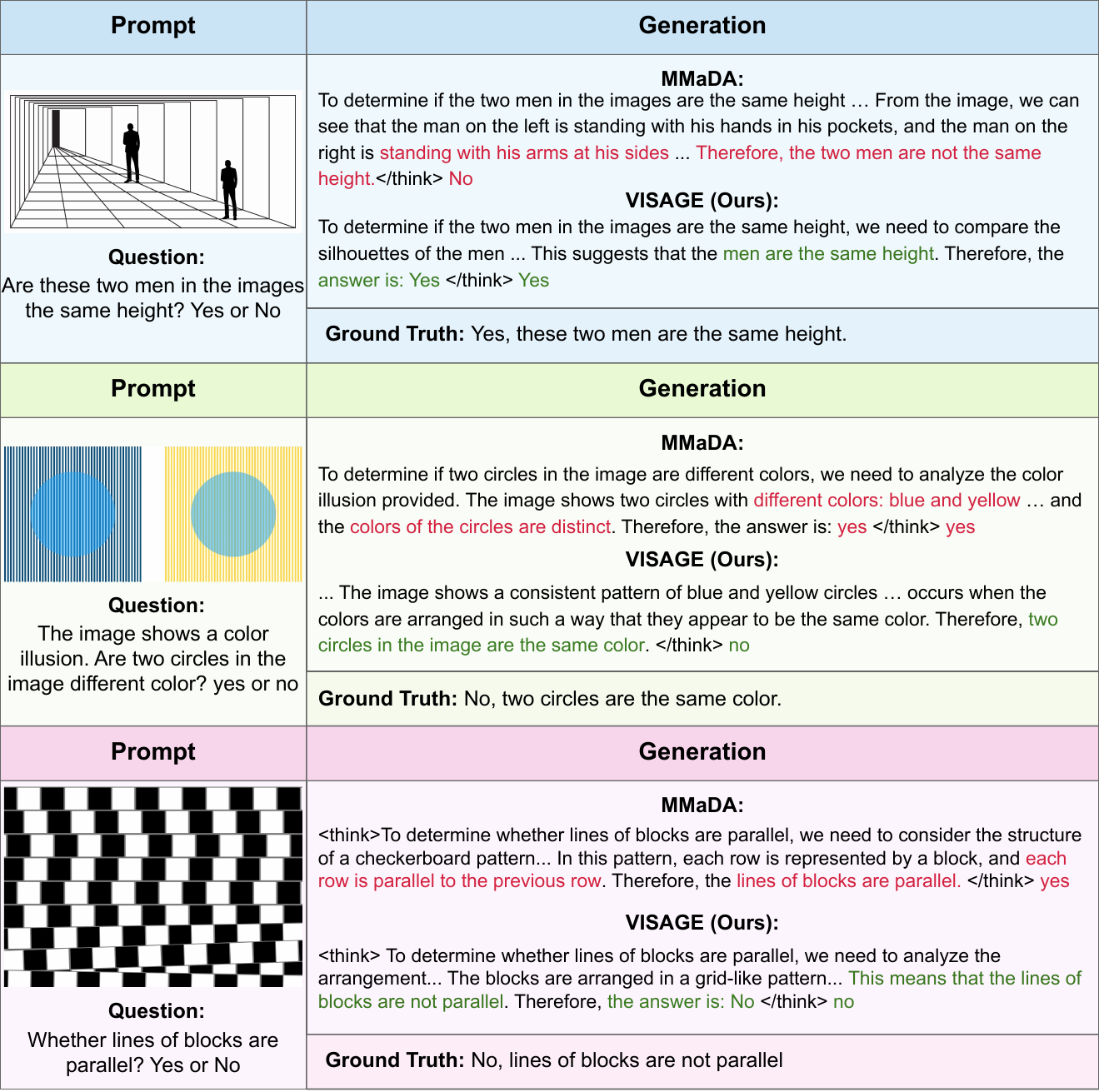}
\caption{
\textbf{Qualitative comparison on HallusionBench.} We present MMaDA (baseline) and VISAGE generations across three deceptive visual contexts. \textit{Top row:} A perspective illusion where MMaDA hallucinates a height difference, while VISAGE correctly anchors its prediction on the actual silhouettes. \textit{Middle row:} A color illusion where the baseline misinterprets uniform circles as distinct colors due to the background patterns. \textit{Bottom row:} A geometric deception where MMaDA relies on a linguistic shortcut about checkerboards, whereas VISAGE correctly grounds its answer in the localized visual arrangement. In all cases, VISAGE successfully aligns with the ground truth by suppressing ungrounded language priors.
}
    \label{fig:supp_hb_qualitative}
    \vspace{-1em}
\end{figure*}

Standard decoding algorithms generate ungrounded predictions when localized visual evidence contradicts strong statistical language priors or human perceptual biases. As illustrated in \Cref{fig:supp_hb_qualitative}, the baseline model (MMaDA \cite{yang2025mmada}) is misled by these contextual priors within the HallusionBench benchmark \cite{guan2024hallusionbench}, whereas VISAGE recovers the multimodal objective by penalizing unverified token commitments. For instance, when evaluating a perspective illusion (top row), MMaDA incorporates the probability mass of the converging lines of a drawn hallway, hallucinating that two identical silhouettes are different heights based on their relative placement in the foreground versus the background. VISAGE bypasses the contextual prior, deriving its probability mass directly from the raw spatial dimensions to affirm the silhouettes are identical in height. Similarly, with a color illusion (middle row), the baseline assigns probability mass to a surrounding striped pattern, incorrectly concluding that two identical circles are different colors. By enforcing the spatial consensus, VISAGE suppresses the language shortcut and establishes that the pixel-level visual evidence for both objects is identical. This over-reliance on language priors also extends to geometric deceptions (bottom row). When asked if the lines of blocks in a distorted pattern are parallel, MMaDA defaults to a textual heuristic defining a checkerboard (``each row is parallel to the previous row'') and answers incorrectly. VISAGE bypasses the linguistic assumption, utilizing the localized visual probability mass of the actual arrangement to identify the non-parallel structure.

\subsection{MMMU-val Benchmark}

\begin{figure*}[t]
    \centering
    \includegraphics[width=\textwidth]{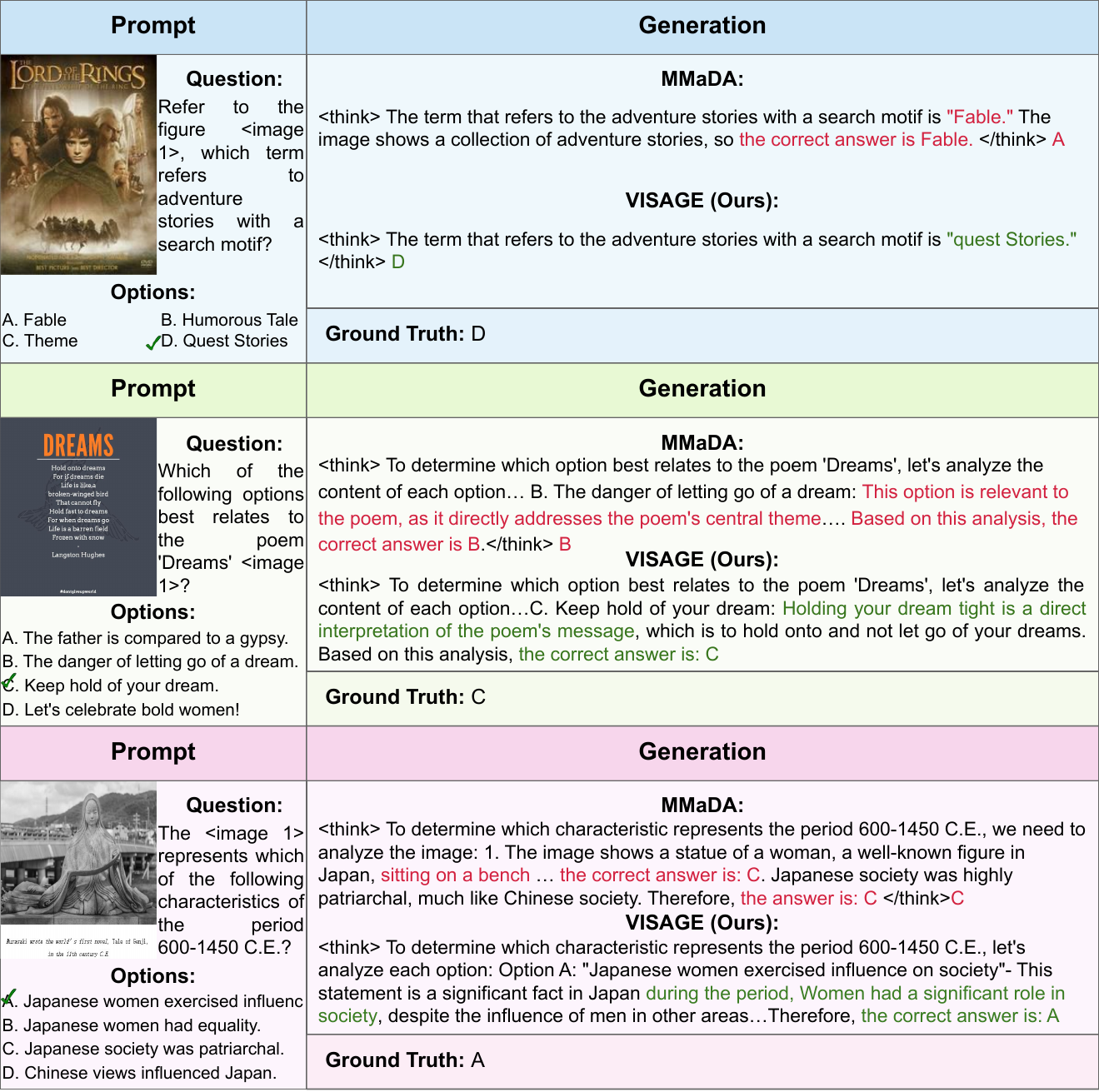}
\caption{
\textbf{Qualitative comparison on MMMU-val.} We present MMaDA (baseline) and VISAGE generations on MMMU-val Benchmark. \textit{Top row:} A visual motif recognition task where MMaDA incorrectly classifies a \textit{Lord of the Rings} poster as a generic ``Fable,'' while VISAGE correctly identifies it as a ``Quest Story''. \textit{Middle row:} A reading comprehension task where the baseline selects a loosely related thematic option for a Langston Hughes poem, whereas VISAGE maintains strict alignment with the text to extract the direct instructional message. \textit{Bottom row:} A historical reasoning task where MMaDA hallucinates a bench and defaults to a broad regional prior (patriarchal society) rather than analyzing the specific visual and contextual cues that indicate Japanese women exercised societal influence during this period. In all three examples, VISAGE successfully anchors its reasoning in the provided evidence, avoiding the shallow linguistic shortcuts that mislead the baseline.
}
    \label{fig:supp_mmmu_qualitative}
    \vspace{-1em}
\end{figure*}

In tasks requiring multi-discipline comprehension, premature token commitments corrupt the intermediate decoding steps of the MDLLM. As demonstrated in \Cref{fig:supp_mmmu_qualitative}, VISAGE stabilizes the refinement process across diverse domains within the MMMU-val benchmark \cite{yue2024mmmu}. In visual motif recognition (top row), MMaDA relies on shallow associations to classify a \textit{Lord of the Rings} poster as a generic ``Fable.'' VISAGE enforces spatial consensus over the visual elements to identify the image as a ``Quest Story.'' This stabilization extends to multimodal reading comprehension; when interpreting Langston Hughes' poem ``Dreams'' (middle row), the baseline model assigns probability mass to a loosely related thematic option. VISAGE maintains alignment with the provided text to extract the direct instructional message. Furthermore, in a historical reasoning scenario (bottom row), MMaDA is skewed by broad statistical priors and object hallucination. When analyzing a Japanese statue, the baseline hallucinates a non-existent ``bench'' and defaults to a generalized regional assumption of a highly patriarchal society. By utilizing localized visual cues, the entropy-based framework suppresses the ungrounded priors, asserting that Japanese women exercised societal influence during the period. Across the tasks, VISAGE anchors its probability mass in the provided evidence, preventing the decoder from committing the ungrounded language shortcuts that confound the baseline.

\subsection{POPE Benchmark}
\label{supp:pope_qual}

\begin{figure*}[t]
    \centering
    \includegraphics[width=\textwidth]{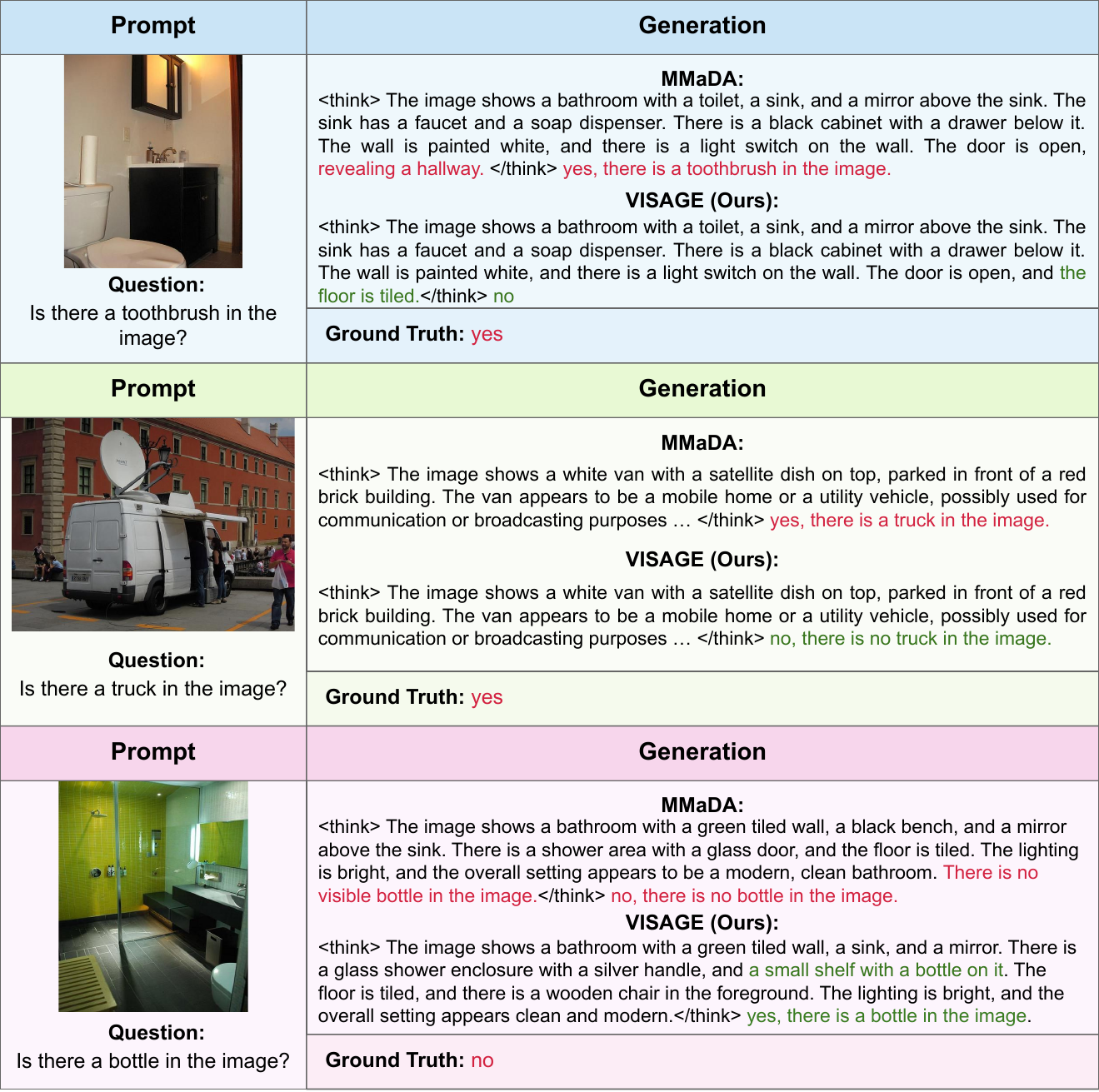}
\caption{
\textbf{Ground-truth label noise in POPE.} We present examples where VISAGE correctly grounds its generation in the visual evidence but is penalized as incorrect due to flawed or missing annotations inherited from the MS COCO dataset. \textit{Top row:} The ground truth falsely asserts a toothbrush is present. MMaDA hallucinates the object to satisfy the prompt, while VISAGE correctly refutes its existence. \textit{Middle row:} The ground truth loosely labels a white van as a ``truck.'' VISAGE precisely identifies the vehicle as a van and rejects the generic truck label, whereas the baseline complies. \textit{Bottom row:} The ground truth fails to account for unannotated bottles resting on the shower shelf. VISAGE successfully discovers them through rigorous spatial grounding. These instances demonstrate issues with POPE's annotation quality, thereby capping the measurable relative gains on the benchmark.
}
    \label{fig:supp_pope_qualitative}
    \vspace{-1em}
\end{figure*}

While VISAGE mitigates hallucinations, the measurable relative gain on the POPE benchmark is capped by inherited label noise. As illustrated in \Cref{fig:supp_pope_qualitative}, POPE \cite{li2023evaluating} evaluation relies on MS COCO annotations, which contain omissions and imprecise labels \cite{tkachenko2023objectlab}. In these scenarios, the spatial entropy consensus selected by VISAGE conflicts with the noisy ground truth. For instance, when the ground truth asserts the presence of a toothbrush in a bathroom scene (top row), MMaDA hallucinates the object to satisfy the affirmative prompt. Because VISAGE demands concentrated visual probability mass, the framework refutes the object's existence but is penalized as a false negative. Similarly, when the dataset loosely categorizes a white utility van as a ``truck'' (middle row), the baseline complies with the prompt. The re-ranking framework identifies the vehicle as a van and rejects the generic label, resulting in a dataset penalty. Furthermore, the spatial consensus allows VISAGE to discover valid, unannotated objects that the baseline misses, such as a bottle resting on a shower shelf (bottom row). These instances highlight that while confidence-based models overfit to flawed dataset priors, VISAGE requires token commitments to exhibit localized visual evidence.

\subsection{Boundary Conditions: Temporal Dynamics}
\label{supp:temporal_failure}

\begin{figure*}[t]
    \centering
    \includegraphics[width=0.9\textwidth]{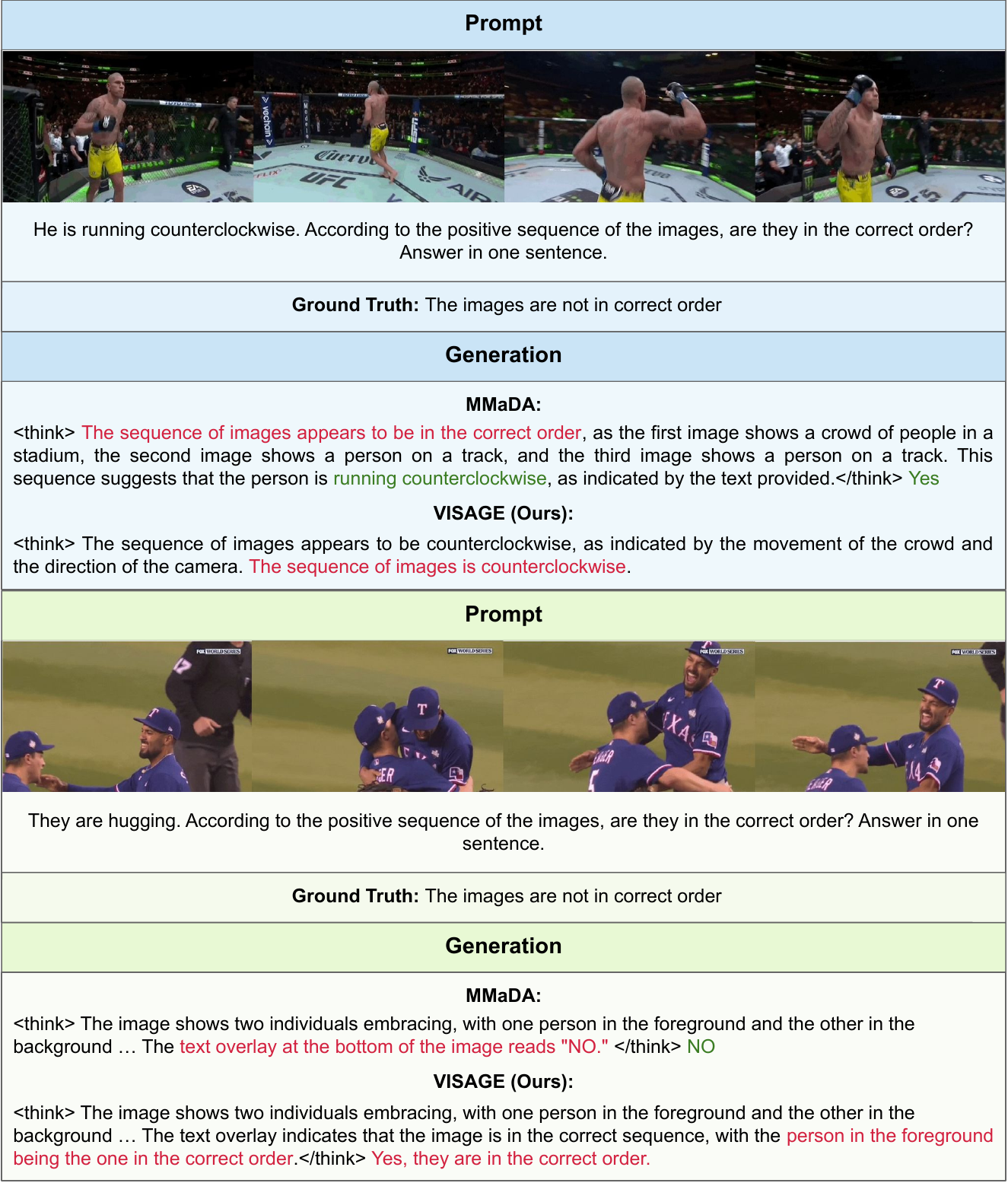}

\caption{
\textbf{Temporal failure cases.} We present examples where both the baseline (MMaDA) and VISAGE struggle to evaluate out-of-order video frames. \textit{Top row:} Given a shuffled sequence of a fighter moving around in the fighting ring, both models erroneously confirm the chronological order, failing to capture the true motion dynamics. However, the baseline changes it's answer to ``Yes'' in the end. \textit{Bottom row:} For a shuffled sequence of baseball players hugging, VISAGE again fails to identify the temporal inconsistency. Notably, while the baseline outputs the correct final answer (``NO''), its intermediate reasoning reveals hallucinations. It falsely claims a non-existent text overlay that provided the answer. These cases demonstrate that without explicit spatio-temporal modeling, both methods remain vulnerable to temporal reasoning failures, even when the baseline occasionally outputs the correct label by chance.
}
    \label{fig:supp_failure_qualitative}
    \vspace{-1em}
\end{figure*}

Because VISAGE derives the re-ranking penalty strictly from spatial attention distributions, the framework inherits the temporal boundaries of the underlying MDLLM. As shown in \Cref{fig:supp_failure_qualitative}, evaluating temporally shuffled video sequences demonstrates this architectural boundary. When presented with out-of-order frames of a fighter moving around in a fighting ring (top row), both models lack a mechanism to verify chronological ordering. The baseline exhibits unstable decoding steps, shifting the output to an incorrect affirmative response. This shared limitation extends to a shuffled sequence of baseball players embracing (bottom row). Because the image-level spatial entropy remains localized on the subjects in each independent frame, VISAGE validates the spatial presence of the actors but does not penalize the sequence disorder. While the baseline outputs the correct final label (``NO'') for the prompt, the intermediate decoding steps rely on hallucinations, generating a non-existent ``NO'' text overlay to justify the decision. These instances confirm that while entropy-based re-ranking enforces strict spatial localization, integrating cross-frame temporal dynamics represents a structural boundary for current image-to-text parallel decoding frameworks.

\section{LLM-as-a-judge Evaluation}
\label{supp:prompts}
To evaluate the reasoning text generated by the MDLLM, we employ an LLM-as-a-judge framework with prompts tailored to each benchmark. For HallusionBench we adopt the evaluation prompt from \cite{guan2024hallusionbench} (P1). The evaluator acts as an ``intelligent teacher'' to assess whether the predicted answer conflicts with the ground truth, classifying the response as ``correct,'' ``incorrect,'' or ``unclear.'' Conversely, the diverse multi-discipline questions in the MMMU benchmark require a targeted evaluator approach, for which we utilize the prompt design introduced in \cite{hong2026widein}. Because the MDLLM generates lengthy intermediate decoding steps, the P2 prompt instructs the judge to isolate the final answer and assign a binary score (1 or 0). The prompt is designed to handle multiple-choice and open-ended formats by accounting for semantic equivalence, numerical precision, and correct units, while ignoring superficial formatting differences.

\begin{tcolorbox}[
    title=P1: HallusionBench LLM-as-a-judge Prompt,
    breakable,
    enhanced jigsaw,
    colback=pbBack,
    colframe=pbFrame,
    coltitle=pbAccent,
    colbacktitle=pbHeader,
    fonttitle=\bfseries,
    leftrule=0.5mm,
    rightrule=0.5mm,
    toprule=0.5mm,
    bottomrule=0.5mm,
    top=2mm,
    bottom=2mm,
    arc=2mm,
    boxsep=4pt,
    drop shadow
]
Imagine you are an intelligent teacher. Thoroughly read the question, reference answer and the prediction answer to ensure a clear understanding of the information provided. Assess the correctness of the predictions. If the prediction answer does not conflict with the reference answer, please generate “correct”. If the prediction answer conflict with the reference answer, please generate “incorrect”. If the prediction answer is unclear about the answer, please generate "unclear".
\end{tcolorbox}

\begin{tcolorbox}[
    title=P2: MMMU LLM-as-a-judge Prompt,
    breakable,
    enhanced jigsaw,
    colback=pbBack,
    colframe=pbFrame,
    coltitle=pbAccent,
    colbacktitle=pbHeader,
    fonttitle=\bfseries,
    leftrule=0.5mm,
    rightrule=0.5mm,
    toprule=0.5mm,
    bottomrule=0.5mm,
    top=2mm,
    bottom=2mm,
    arc=2mm,
    boxsep=4pt,
    drop shadow
]
You are a strict evaluator assessing answer correctness. You must output 1 for fully correct answers and 0 for any other case.\\
\# Input\\
Question: \{question\}\\
Ground Truth Answer: \{answer\}\\
Model Prediction: \{pred\}\\

\# Evaluation Rules\\
- The model prediction may contain the reasoning process, you should spot the final answer from it.\\
- For multiple-choice questions: Score 1 if the predicted answer matches the ground truth answer, it can be directly in option letters or the content of the options.\\
- For open-ended questions:\\
  \quad* Score 1 if the prediction matches the answer semantically, it can be in different format.\\
  \quad* Score 0 for partially correct answers or answers with extra incorrect information, even if the reasoning process is correct.\\
- Ignore minor differences in formatting, capitalization, or spacing since the model may explain in a different way.\\
- Treat numerical answers as correct if they match within reasonable precision\\
- For questions requiring units, both value and unit must be correct\\

\# Strict Output format
0 or 1
\end{tcolorbox}

\end{document}